\documentclass[conference]{IEEEtran}
\IEEEoverridecommandlockouts

\usepackage{amssymb}
\usepackage{booktabs}
\usepackage{tabularx}
\usepackage{makecell}
\usepackage{multirow} 
\usepackage{graphicx} 
\usepackage{url}
\usepackage{enumitem}  
\usepackage{tikz}
\usepackage{natbib}
\usepackage{pgfplots} 
\usepackage{booktabs}
\usepackage{multirow}
\usepackage{graphicx}
\usepackage{xcolor}
\usepackage{makecell}
\usepackage{array}
\usepackage{makecell}
\usepackage{capt-of}
\usepackage{cuted}
\usepackage[font=small, skip=2pt]{caption}
\usepackage{capt-of}
\definecolor{darkgreen}{RGB}{0,120,0}
\usepackage{xcolor}
\usepackage{hyperref}
\hypersetup{
    colorlinks=false,
    citebordercolor={0 1 0},  
    linkbordercolor={0 0 1},  
    urlbordercolor={1 0 0}    
}
\begin{document}

\title{Towards Enhancing 3D Spatial Reasoning in Medical Multimodal Large Language Models}

\author{
    \IEEEauthorblockN{
        Zhuoyuan Fu\textsuperscript{1,$\dagger$}, 
        Zeshang Li\textsuperscript{1,$\dagger$}, 
        Yiqiong Zhang\textsuperscript{1,$\dagger$},  
        Hangui Lin\textsuperscript{1}, 
        Yan Shu\textsuperscript{2}, 
        Yan Li\textsuperscript{1,*}, 
        Binyang Li\textsuperscript{1}, 
        Yaru Zhao\textsuperscript{1}
    }
    \vspace{0.15cm} 
    \IEEEauthorblockA{
        \textsuperscript{1}\textit{University of International Relations, Beijing, China} \\
        \textsuperscript{2}\textit{University of Trento, Trento, Italy}
    }
    
    \thanks{$^{\dagger}$ These authors contributed equally to this work.}
    \thanks{* Corresponding author: liyan@uir.edu.cn. This work was supported in part by the Fundamental Research Funds for the Central Universities under Grant 3262025T82, the Beijing Natural Science Foundation under Grant 4262075, and the Research Funds for NSD Construction, University of International Relations, under Grant 3262026T23.}
}

\maketitle


\begin{abstract}
While Multimodal Large Language Models (MLLMs) have demonstrated remarkable success in 2D medical image understanding, their extension to 3D volumetric imaging remains hindered by prohibitive annotation costs and dataset opacity. Current data formats, predominantly consisting of rigid Visual Question Answering (VQA) pairs or unstructured final clinical reports, typically fail to capture explicit clinical reasoning. To address this limitation, we introduce a large-scale structured reasoning dataset constructed via a novel slice-wise data synthesis paradigm. Inspired by the genuine diagnostic workflow of radiologists, this paradigm models visual cognition by decomposing the complex 3D reading process, translating global clinical priors into fine-grained, per-slice observations that are subsequently synthesized into an interpretable Chain-of-Thought (CoT). Crucially, this synthesized reasoning framework enforces essential clinical principles: sequential spatial tracking, multi-slice spatial awareness for artifact mitigation, and differential exclusion. To validate this approach, we instruction-tune a standard 2D-pretrained MLLM baseline using the synthesized data to enhance its volumetric comprehension. Comprehensive evaluations across multiple 3D medical benchmarks demonstrate that our method yields significant performance improvements over the 2D baseline. Furthermore, the resulting model exhibits robust spatial reasoning capabilities and rivals resource-intensive native 3D architectures, effectively bridging the performance gap. Ultimately, this data-centric strategy unlocks deep volumetric understanding and highly interpretable clinical logic without requiring computationally expensive 3D-specific pre-training. \textbf{The complete repository, including datasets and training workflows, is publicly available at \url{https://github.com/2020420145009/hounsfield}.}
\end{abstract}


\IEEEpeerreviewmaketitle

\section{Introduction}
\label{sec:intro}

\begin{figure*}[t] 
    \centering
    \includegraphics[width=\textwidth]{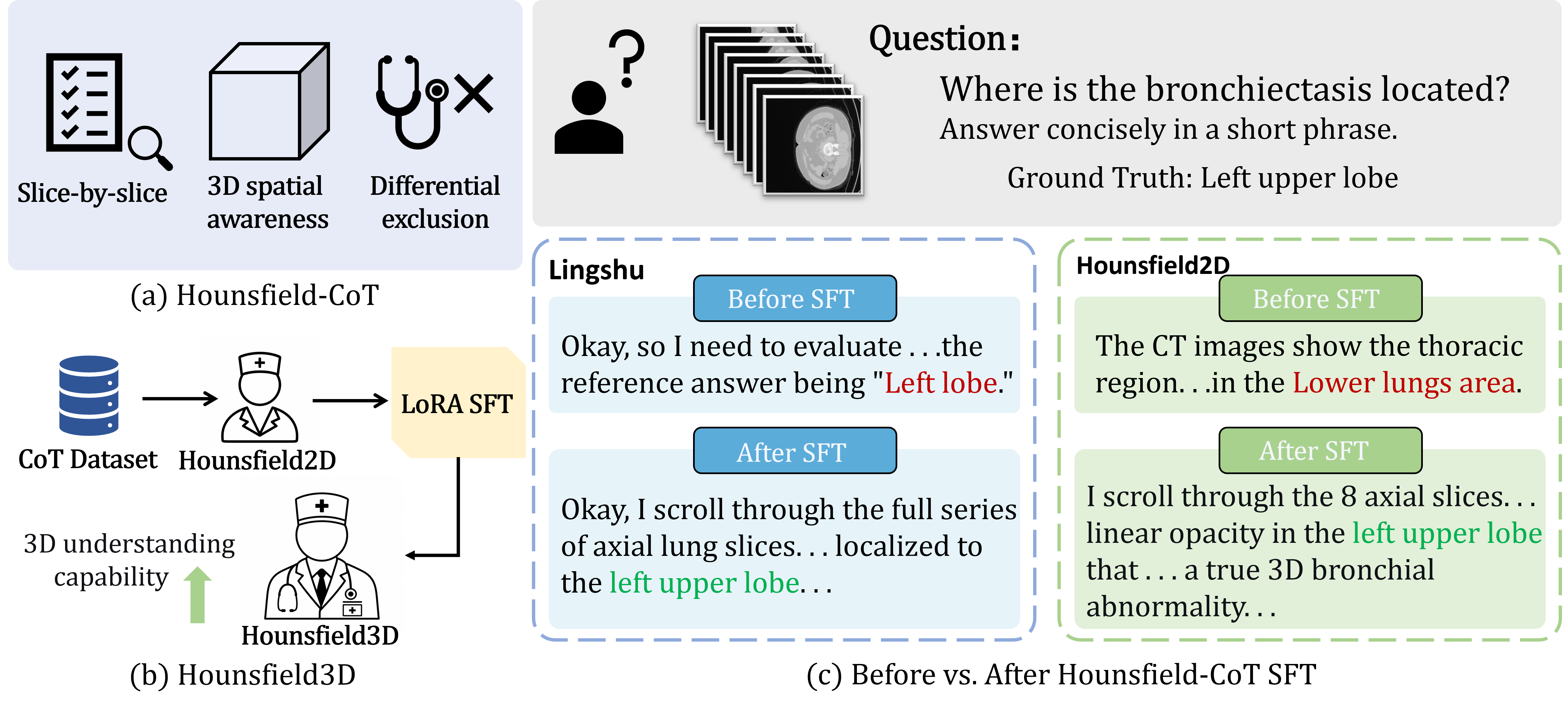}
    \caption{\textbf{Overview of Hounsfield-CoT and model performance refinement.} (a) Our paradigm integrates sequential slice-by-slice tracking, 3D spatial awareness, and differential exclusion as core competencies. (b) We validate this framework by instruction-tuning the Hounsfield2D model on our curated CoT dataset via LoRA to acquire advanced volumetric reasoning capabilities, resulting in the Hounsfield-3D model. (c) Qualitative comparisons before and after SFT demonstrate that while baseline models hallucinate spatial coordinates due to opaque black-box inference, our paradigm enables explicit volumetric reasoning that successfully resolves complex localization queries.}
    \label{fig:teaser_comparison}
\end{figure*}

Multimodal large language models (MLLMs)~\cite{yu2025umit,alsaad2024multimodal,tu2024towards} have been increasingly adapted to the medical domain~\cite{xu2025lingshu,shu2025fleming,jiang2025hulu} by leveraging large-scale 2D image-text pairs~\cite{hamamci2024developing,li2025towards}. While these models demonstrate competitive performance on standard 2D tasks~\cite{li2023llava,zhang2023biomedclip}, clinical practice is fundamentally grounded in 3D modalities (e.g., CT and MRI)~\cite{blankemeier2024merlin,he2025vista3d}. Recent attempts to extend MLLMs to 3D imaging~\cite{bai2024m3d,mazurowski2023segment,shi2025med} have primarily focused on representation learning and cross-modal alignment. Consequently, complex, multi-step reasoning over volumetric data remains largely unaddressed~\cite{zhou2026beyond,sambara20253dreasonknee}.

Chain-of-thought (CoT) prompting~\cite{wang2025multimodal} has proven highly effective in structuring intermediate reasoning steps for complex medical tasks~\cite{liu2026clincot,wei2024mc,liu2025fleming}. However, current medical CoT methodologies are predominantly confined to 2D or text-based scenarios~\cite{jiang2026m3cotbench}. While recent efforts have explored shared grounded reasoning interfaces to facilitate 2D-to-3D transfer~\cite{chen2026unireason}, directly generalizing these approaches to native 3D volumes remains non-trivial. Real-world clinical volumetric diagnosis inherently relies on a sequential, slice-by-slice reading workflow to track lesion evolution and integrate partial cross-sectional views into a coherent 3D anatomical understanding. Existing datasets fail to capture this slice-wise cognitive process, frequently reducing 3D volumes to collapsed 2D projections or treating them as unstructured black boxes. This critical absence of spatially-grounded CoT supervision leaves volumetric reasoning models without clinically aligned, interpretable guidance.

To address these challenges, we propose a novel slice-wise data synthesis paradigm that explicitly models the visual cognition of expert radiologists. Instead of relying on holistic volume-level summaries, we employ a dual-agent pipeline to simulate the clinical diagnostic workflow: an \textit{Observer} agent first extracts fine-grained, per-slice findings from global clinical reports, and a \textit{Synthesizer} agent then integrates these discrete observations into a structured, spatially-grounded CoT. Crucially, every generated reasoning chain is rigorously constrained to enforce three core clinical principles: sequential spatial tracking, true 3D spatial awareness, and differential exclusion. Driven by this paradigm, we introduce \textbf{Hounsfield-CoT}, a large-scale 3D medical reasoning dataset providing explicit supervision for volumetric understanding. Derived exclusively from the CT-RATE dataset, Hounsfield-CoT comprises 11.2k rigorously curated instances targeting foundational spatial reasoning. As illustrated in Fig.~\ref{fig:teaser_comparison}, this data-centric approach successfully replaces opaque reasoning shortcuts with highly interpretable clinical logic.

By instruction-tuning a 2D-pretrained MLLM on Hounsfield-CoT, we bridge the semantic gap between 2D pre-training and complex volumetric understanding. At inference time, the adapted model systematically leverages the learned slice-wise cognitive pattern to observe, track, and synthesize findings, yielding significantly stronger and more transparent diagnostic performance.

In summary, our main contributions are three-fold:

\begin{enumerate}[nosep,leftmargin=*]
\item \textbf{We propose a novel slice-wise data synthesis paradigm for 3D medical reasoning.} Simulating the diagnostic workflow of expert radiologists, we introduce a dual-agent (Observer-Synthesizer) framework. This paradigm explicitly translates global volume-level priors into fine-grained, per-slice observations, rigorously enforcing sequential spatial tracking, true 3D spatial awareness, and differential exclusion.

\item \textbf{We introduce Hounsfield-CoT, a large-scale, spatially-grounded reasoning dataset.} Driven by our synthesis paradigm, we curate 11.2k high-quality 3D reasoning instances. This dataset bridges the critical gap in explicit volumetric supervision, replacing opaque reasoning shortcuts with transparent, step-by-step clinical logic.

\item \textbf{We establish a cost-effective 2D-to-3D reasoning transfer framework and demonstrate superior diagnostic performance.} By instruction-tuning a standard 2D-pretrained MLLM using Hounsfield-CoT, we effectively unlock its latent 3D spatial reasoning capabilities without computationally prohibitive 3D-specific pre-training. Extensive evaluations reveal substantial performance gains alongside highly interpretable diagnostic rationales, successfully mitigating partial-volume artifacts and spatial hallucinations.
\end{enumerate}

\section{Related Work}

\subsection{Traditional Paradigms in Medical MLLMs}
\label{subsec:dimensional_paradigms}

While Medical MLLMs have achieved remarkable success in 2D imaging, extending these capabilities to 3D volumetric data introduces significant computational and architectural challenges. Current research predominantly diverges into two paradigms: native 3D modeling and 2D-to-3D adaptation. Pioneering explicit 3D models—such as M3D~\cite{bai2024m3d} for unified multi-task learning, LLaVA-3D~\cite{zhu2024llava} for visual grounding, and Med3DVLM~\cite{xin2025med3dvlm} for efficient vision-language alignment—directly process entire volumes to capture holistic anatomical contexts. However, these native approaches demand massive computational resources and operate as opaque ``black boxes'' lacking diagnostic transparency.

Alternatively, 2.5D feature fusion strategies bypass these bottlenecks by leveraging mature 2D backbones. Med-2E3~\cite{shi2025med} aggregates 2D slice embeddings to reduce training costs, while BrainGPT~\cite{li2025towards} optimizes sequence-based slice processing for neuroimaging. Though efficient, these methods inherently risk losing vital inter-slice continuity. To mitigate the opacity of 3D models and the spatial reasoning gaps of 2.5D methods, recent works have explored CoT reasoning. Methods like MC-CoT~\cite{wei2024mc}, MedCoT~\cite{liu2024medcot}, and the guideline-aligned ClinCoT~\cite{liu2026clincot} successfully elicit step-by-step rationales to enhance transparency. Unfortunately, these CoT approaches remain almost entirely confined to 2D modalities or rely on rigid, static templates. Although nascent efforts attempt to bridge this divide via shared grounded reasoning interfaces for 2D-to-3D transfer~\cite{chen2026unireason}, they do not fully address the complexities of native volumetric logic. Furthermore, as diagnostic reasoning chains extend across 3D space, maintaining dynamic cross-modal coordination becomes imperative to prevent attention collapse and visual-textual misalignment~\cite{lin2026dyco}. Consequently, there remains a critical gap in adaptive, transparent reasoning capable of explicitly unrolling diagnostic logic across complex 3D volumes.
\begin{figure*}[t]
    \centering
    \includegraphics[width=\textwidth]{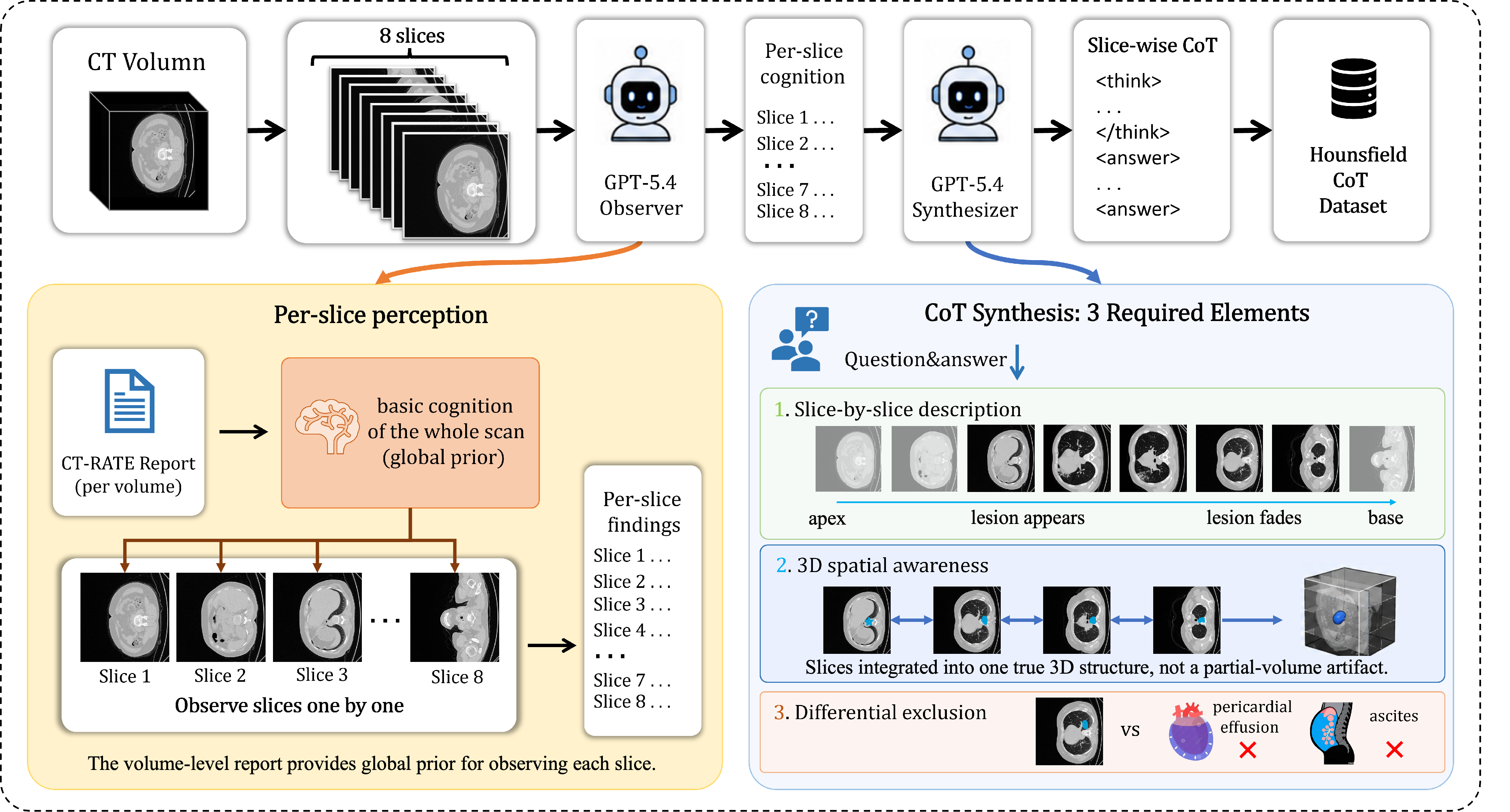} 
    \caption{\textbf{Overview of the Hounsfield-CoT data synthesis pipeline.} The workflow explicitly models the slice-by-slice reading process of radiologists to transform raw 3D CT volumes into a high-quality CoT dataset, entirely bypassing the need for bounding boxes or image cropping. \textbf{Top: Dual-Agent Pipeline.} A raw CT volume is sampled into 8 slices and processed sequentially by a GPT-5.4 Observer and a GPT-5.4 Synthesizer, ultimately generating a dataset strictly formatted with \texttt{<think>} and \texttt{<answer>} blocks. \textbf{Bottom Left: Per-slice Perception.} Guided by the volume-level CT-RATE clinical report as a global prior, the Observer agent examines the scan slice-by-slice, extracting fine-grained, discrete per-slice findings. \textbf{Bottom Right: CoT Synthesis.} The Synthesizer integrates these discrete observations into a coherent reasoning chain that strictly mandates three required clinical elements: (1) sequential slice-by-slice description (e.g., tracking a lesion from appearance to fading, where the target sequence is visually decoupled via opacity dimming to maintain the bounding-box-free constraint); (2) true 3D spatial awareness (integrating slices to differentiate genuine 3D structures from partial-volume artifacts); and (3) differential exclusion against competing clinical diagnoses (explicitly ruling out confounding conditions).}
    \label{fig:curation_pipeline}
    \vspace{-1.5em}
\end{figure*}

\subsection{Training Datasets for Medical MLLMs}
Data synthesis has emerged as a critical strategy to overcome the severe scarcity of high-quality medical multi-modal training data. For standard 2D imaging, scalable instruction-following datasets have been successfully constructed through diverse pipelines. For instance, PMC-VQA~\cite{zhang2024pmc} leverages massive biomedical literature mining to extract extensive figure-caption pairs, while MedTrinity-25M~\cite{xie2024medtrinity} utilizes retrieval-augmented generation to provide multigranular spatial annotations. Furthermore, works like BioMed-VITAL~\cite{cui2024biomedical} and HuatuoGPT-Vision~\cite{chen2024towards} refine these datasets by aligning model outputs directly with human clinician preferences and real-world diagnostic behaviors. 

Transitioning to the 3D domain, pioneering efforts have begun scaling volumetric corpora. M3D~\cite{bai2024m3d} constructs extensive 3D VQA pairs for unified multi-task learning, and RadGenome-Chest CT~\cite{zhang2024radgenome} advances the field by linking diagnostic text to specific spatial anatomical regions to generate grounded reports. 

Nevertheless, a critical limitation persists across these foundational efforts: existing 3D datasets predominantly exist as flat, end-to-end question-answer pairs or static final reports. This macroscopic format structurally conceals the intermediate, slice-by-slice analytical processes and differential exclusion strategies naturally employed by physicians. Consequently, models trained on such data are prone to shortcut learning rather than genuine spatial understanding. This underscores the urgent need for a novel data synthesis paradigm that explicitly unrolls and embeds dynamic, multi-step reasoning trajectories into 3D clinical datasets.


\section{Dataset Construction}
\label{sec:methodology}

In this section, we detail the construction of the Hounsfield-CoT dataset, specifically engineered to bridge the semantic gap between raw 3D medical volumes and expert-level diagnostic reasoning. To address core challenges in 3D clinical data—namely the opacity of end-to-end models and the absence of explicit, spatially grounded reasoning chains—we implemented a systematic, data-centric curation pipeline (illustrated in Fig.~\ref{fig:curation_pipeline}). Moving away from static multi-source aggregation, we focused exclusively on the large-scale CT-RATE dataset to construct high-quality, task-specific reasoning trajectories for 3D-RadVQA. Central to our approach is a novel \textbf{Observer-Synthesizer Dual-Agent Framework}, which explicitly mimics the clinical slice-by-slice reading workflow. This systematic approach effectively transforms traditionally opaque visual question-answer pairs into a transparent, clinically interpretable corpus comprising 11.2k strictly curated instances.

\subsection{Raw Data Preparation}
\label{subsec:data_preparation}

The objective of this initial phase is to establish a robust factual anchor from the raw CT-RATE dataset and align it with standardized evaluation formats prior to LLM-based reasoning synthesis.

Unlike conventional 2D VQA datasets where a single image maps to a brief answer, 3D volumetric data contains dense anatomical structures. To construct a reliable factual basis, we extract the corresponding volume-level clinical reports for each scan. These reports, written by expert radiologists, encapsulate the ground-truth pathological findings of the entire 3D volume and serve as an indispensable global prior for our synthesis pipeline. 

Simultaneously, we restrict our task formulation to the core competencies evaluated in the 3D-RadVQA benchmark: Anomaly Detection (T1), Image Observation (T2), Medical Computation (T3), and Existence Detection (T4). For each scan, we structure the ground-truth metadata to automatically instantiate corresponding queries in both Open-ended and Multiple-Choice Question (MCQ) formats. This rigorous formulation ensures that the generated dataset directly targets foundational 3D spatial reasoning capabilities.

\begin{figure*}[t]
    \centering
    \includegraphics[width=\textwidth]{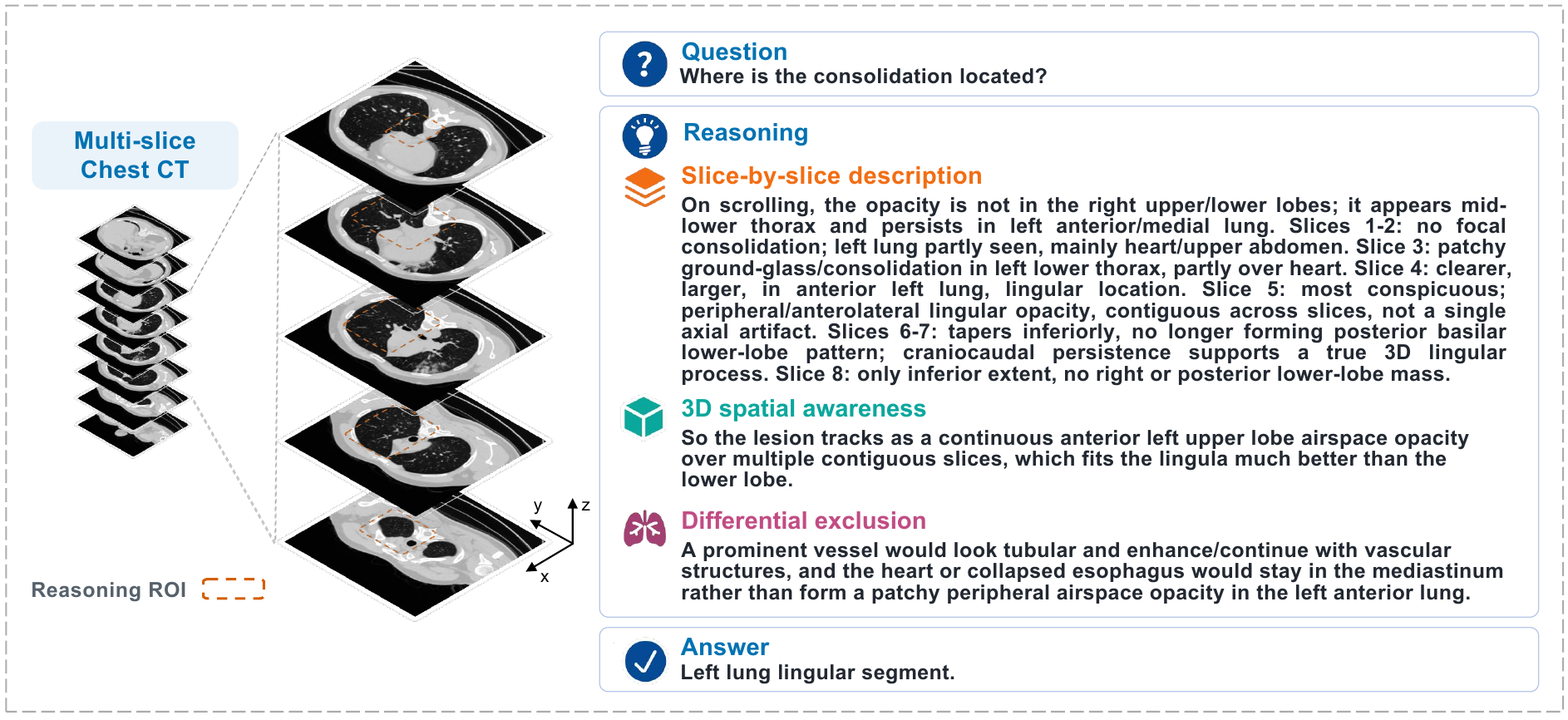}

    \caption{\textbf{Qualitative example of the generated Hounsfield-CoT reasoning path.} The presented case strictly adheres to our proposed three required clinical elements. In this consolidation localization task, the model provides a \textit{Sequential slice-by-slice description} of the abnormal opacity, demonstrates \textit{True 3D spatial awareness} by integrating the lingular lesion across contiguous slices, and performs \textit{Differential exclusion} to distinguish it from vascular structures, mediastinal anatomy, or single-axial partial-volume artifacts. The reasoning logic is fully unrolled prior to final answer formulation.}

    \label{fig:qualitative_cases}
    \vspace{-1.5em}
\end{figure*}

\subsection{Slice-wise CoT Synthesis via Dual-Agent Framework}
\label{subsec:instruction_synthesis}

Following data preparation, this phase focuses on generating explicit, spatially grounded reasoning trajectories. Previously, vision-language models tended to act as ``end-to-end'' black boxes, outputting final diagnostic results without transparent reasoning processes. Directly applying standard CoT to 3D images often fails, as volumetric data introduces complex spatial variations (e.g., partial-volume artifacts) that language priors alone cannot resolve. 

To overcome this, we draw inspiration from the systematic diagnostic patterns of professional radiologists and propose a \textbf{Dual-Agent Synthesis Framework} powered by a Large Language Model (e.g., GPT-5.4). To ensure the synthesized CoT is a clinically valid deduction rather than a linguistic hallucination, the generation process is meticulously controlled and mandated to manifest three core cognitive elements: (1) \textbf{Sequential Spatial Tracking:} Documenting the appearance, peak manifestation, and fading of the target structure across consecutive Z-axis slices. (2) \textbf{True 3D Spatial Awareness:} Integrating discrete 2D observations to explicitly differentiate genuine 3D anatomical abnormalities from localized partial-volume artifacts. (3) \textbf{Differential Exclusion:} Conducting a comparative analysis against standard morphology to exclude clinically similar competing diagnoses. Concrete qualitative examples of these elements are provided in Fig.~\ref{fig:qualitative_cases}.

We operationalize this cognitive framework through a cohesive pipeline that explicitly decouples observation from synthesis. During \textbf{Per-slice Perception}, an \textit{Observer} agent utilizes the volume-level clinical report as a global prior to sequentially examine the scan and extract discrete, localized findings. Subsequently, in the \textbf{CoT Synthesis} phase, a \textit{Synthesizer} agent integrates these discrete observations into a continuous reasoning chain strictly governed by the three aforementioned clinical elements. 

To ensure structural consistency, the generated reasoning trajectories are verified and decoupled, with the reasoning process confined within \texttt{<think>} tags and the final diagnosis in \texttt{<answer>} tags. Ultimately, this streamlined process yields the final Hounsfield-CoT dataset of 11.2k high-quality instances. This dense, highly interpretable cognitive foundation provides a distinct advantage for subsequent instruction tuning, enabling MLLMs to acquire robust volumetric understanding without the prohibitive cost of 3D-specific pre-training. Such a paradigm of utilizing explicitly grounded reasoning trajectories to unlock domain-specific capabilities closely mirrors recent successful dataset innovations in other complex spatial modalities, such as pixel-grounded reasoning for earth observation~\cite{shu2026terrascope}.

\subsection{Dataset Statistics and Characteristics}
\label{subsec:dataset_statistics}

The finalized Hounsfield-CoT dataset is systematically structured to cover the four core competencies of 3D-RadVQA. As detailed in Table~\ref{tab:dataset_stats}, the dataset maintains a balanced format distribution: 51.2\% of the data (T1--T3) consists of open-ended generative queries, while the remaining 48.8\% (T4) is dedicated to yes-no multiple-choice verification, supporting comprehensive evaluation protocols.

A defining characteristic of our dataset is its dense, highly interpretable reasoning structure. When processing the volumetric scans (typically featuring a mode of 8 slices), the synthesized \texttt{<think>} blocks explicitly unroll complex spatial tracking and differential diagnostic processes. This yields a striking average token length ratio of 92.9:1 between the detailed reasoning chains and the highly concise final \texttt{<answer>} outputs. Such a stark contrast underscores our primary objective: shifting MLLMs from superficial vision-to-text mapping toward verifiable, deep volumetric reasoning, thereby unlocking genuine 3D spatial awareness without the prohibitive cost of explicit 3D pre-training.

\begin{table}[t]
    \centering
    \small
    \renewcommand{\arraystretch}{1.15}
    \caption{\textbf{Overview of the Hounsfield-CoT dataset.}}
    \label{tab:dataset_stats}

    \begin{tabular*}{0.82\linewidth}{@{\extracolsep{\fill}}l r@{}}
        \toprule
        \multicolumn{2}{c}{\textit{Dataset Overview}} \\
        \midrule
        Avg. \texttt{\textless think\textgreater} tokens  & 255.7 \\
        Avg. \texttt{\textless answer\textgreater} tokens & 2.8 \\
        \midrule
        \multicolumn{2}{c}{\textit{Instance Taxonomy}} \\
        \midrule
        Anomaly Detection (T1)    & 1,711 \\
        Image Observation (T2)    & 1,950 \\
        Medical Computation (T3)  & 2,102 \\
        Existence Detection (T4)  & 5,500 \\
        \bottomrule
    \end{tabular*}
\end{table}

\section{Model}
\label{sec:model_and_finetuning}

Our proposed approach, \textbf{Hounsfield3D}, aims to unlock multi-step spatial reasoning for 3D medical imaging by combining lightweight volumetric processing with our structured CoT supervision. Rather than designing a computationally expensive 3D-native architecture from scratch, we adapt a standard 2D-pretrained multimodal baseline to volumetric tasks, demonstrating the efficacy of our data-centric paradigm.

\subsection{Model Architecture and Spatial Serialization}
\label{subsec:architecture_and_processing}
We build upon a robust, standard Vision-Language framework comprising a vision encoder (e.g., \texttt{InternViT}) coupled with a decoder-only LLM (e.g., \texttt{Qwen2}) via an MLP projector. To effectively handle high-dimensional clinical CT data without the massive computational overhead of dense 3D convolutions—and strictly avoiding heuristic pre-processing steps like bounding box (BBox) extraction or image cropping—we employ a straightforward 2.5D spatial serialization strategy.

Specifically, for each CT volume, we extract a sequence of contiguous slices along the z-axis,  comprising eight slices. These slices are independently processed by the 2D vision encoder and sequentially embedded into a continuous visual token sequence. This slice-by-slice representation perfectly mirrors the clinical reading workflow and directly aligns with the ``per-slice perception'' logic generated by our Observer agent. It enables the language model to ground its sequential spatial tracking in the visual tokens, effectively capturing global structural coherence and true 3D spatial awareness from discrete 2D inputs.

\subsection{Instruction Tuning and CoT Alignment}
\label{subsec:alignment_strategy}

To validate our synthesized reasoning data, we fine-tune a 2D-pretrained baseline on the \textbf{Hounsfield-CoT} dataset (11.2k instances) using Low-Rank Adaptation (LoRA). 

The model is trained to decouple its internal analytical process from the final clinical answer. By generating reasoning steps within \texttt{<think>} tags prior to the \texttt{<answer>} block, the model explicitly learns to map clinical findings to structured diagnostic logic—specifically sequential slice tracking, artifact mitigation, and differential exclusion. This alignment strategy demonstrates that high-quality, structured CoT supervision effectively transforms standard 2D backbones into robust 3D volumetric reasoners without requiring architectural modifications.

\begin{table*}[t!]
    \centering
    \small
    \renewcommand{\arraystretch}{1.15}
    \setlength{\tabcolsep}{6pt} 
    \caption{\textbf{Main results on 3D‑RadVQA (T1--T4) and DeepchestVQA.}
    We report BERT‑F1 and Accuracy (\%) for open‑ended T1--T3 and multiple‑choice T4, along with average DeepchestVQA zero‑shot accuracy (\%). Our 11.2k Hounsfield‑CoT dataset consistently improves Accuracy, particularly on 2D‑adapted models. Bold indicates the better value within each base/fine‑tuned pair. M3D and other foundation models serve as reference baselines. \textcolor{darkgreen}{$\uparrow$} and \textcolor{red}{$\downarrow$} denote the absolute percentage point changes in Accuracy after fine‑tuning.}
    \label{tab:preliminary_results}
    \resizebox{\textwidth}{!}{
    \begin{tabular}{l c@{\hspace{2pt}} r c @{\hspace{2pt}}r c @{\hspace{2pt}} r @{\hspace{20pt}} c  cc@{\hspace{15pt}}cc}
        \toprule
        \multirow{3}{*}{Model} & 
        \multicolumn{7}{c}{3D-RadVQA} & 
        \multicolumn{4}{c}{DeepchestVQA Avg. (\%)} \\
        \cmidrule(lr){2-8} \cmidrule(lr){9-12}
        & \multicolumn{2}{c}{T1} & \multicolumn{2}{c}{T2} & \multicolumn{2}{c}{T3} & \multicolumn{1}{c}{T4} & \multirow{2}{*}{Recog.} & \multirow{2}{*}{Med.} & \multirow{2}{*}{Vis.} & \multirow{2}{*}{Overall} \\
        \cmidrule(lr){2-3} \cmidrule(lr){4-5} \cmidrule(lr){6-7} \cmidrule(lr){8-8}
        & BERT‑F1 & Acc & BERT‑F1 & Acc & BERT‑F1 & Acc & Acc & & & & \\
        \midrule
        M3D         & 0.922 & 46.0 & 0.907 & 41.9 & 0.948 & 15.4 & 82.3 & 12.8 & 20.4 & 17.6 & 17.7 \\
        Qwen        & 0.650 & 24.7 & 0.624 & 18.3 & 0.444 & 12.3 & 38.3 & 32.2 & 32.3 & 20.5 & 27.1 \\
        InternVL    & 0.824 & 28.1 & 0.824 & 22.5 & 0.780 & 17.1 & 35.1 & 1.7 & 2.4 & 9.7 & 5.5 \\
        Med3DVLM    & 0.868 & 21.0 & 0.874 & 23.5 & 0.908 & 26.6 & 47.0 & 50.0 & 41.5 & 29.4 & 37.9 \\
        Huatuo      & 0.850 & 32.0 & 0.852 & 27.3 & 0.835 & 22.8 & 38.2 & 55.6 & 41.2 & 27.4 & 38.0 \\
        \midrule
        Lingshu (Base)        & \textbf{0.837} & 9.9  & \textbf{0.837} & 14.3 & \textbf{0.879} & 7.8  & 72.4 & \textbf{50.0} & \textbf{40.5} & 22.0 & 34.5 \\
        Lingshu (Ours, FT)    & 0.834 & \textbf{19.4}\rlap{\scriptsize{\,\textcolor{darkgreen}{$\uparrow$9.5}}}  & 0.831 & \textbf{16.1}\rlap{\scriptsize{\,\textcolor{darkgreen}{$\uparrow$1.8}}}  & 0.831 & \textbf{18.6}\rlap{\scriptsize{\,\textcolor{darkgreen}{$\uparrow$10.8}}} & \textbf{76.4}\rlap{\scriptsize{\,\textcolor{darkgreen}{$\uparrow$4.0}}} & 48.3\rlap{\scriptsize{\,\textcolor{red}{$\downarrow$1.7}}} & 40.2\rlap{\scriptsize{\,\textcolor{red}{$\downarrow$0.3}}} & \textbf{30.4}\rlap{\scriptsize{\,\textcolor{darkgreen}{$\uparrow$8.4}}} & \textbf{37.6}\rlap{\scriptsize{\,\textcolor{darkgreen}{$\uparrow$3.1}}} \\
        \midrule
        Hounsfield2D (Base)     & 0.841 & 7.2  & 0.841 & 7.7  & 0.846 & 3.2  & 77.8 & \textbf{51.1} & 39.7 & 26.6 & 36.4 \\
        Hounsfield2D (Ours, FT) & \textbf{0.881} & \textbf{33.5}\rlap{\scriptsize{\,\textcolor{darkgreen}{$\uparrow$26.2}}} & \textbf{0.872} & \textbf{24.1}\rlap{\scriptsize{\,\textcolor{darkgreen}{$\uparrow$16.4}}} & \textbf{0.910} & \textbf{10.3}\rlap{\scriptsize{\,\textcolor{darkgreen}{$\uparrow$7.1}}} & \textbf{80.2}\rlap{\scriptsize{\,\textcolor{darkgreen}{$\uparrow$2.4}}} & 50.6\rlap{\scriptsize{\,\textcolor{red}{$\downarrow$0.5}}} & \textbf{45.1}\rlap{\scriptsize{\,\textcolor{darkgreen}{$\uparrow$5.4}}} & \textbf{29.9}\rlap{\scriptsize{\,\textcolor{darkgreen}{$\uparrow$3.3}}} & \textbf{39.6}\rlap{\scriptsize{\,\textcolor{darkgreen}{$\uparrow$3.2}}} \\
        \bottomrule
    \end{tabular}
    }
    \vspace{-1.5em}
\end{table*}


\section{Experiments}
\label{sec:experiments}
This section empirically evaluates the \textbf{Hounsfield-CoT} paradigm. Our evaluation features rigorous cross-model and cross-benchmark analyses to validate robustness and generalizability, alongside a data scaling ablation to uncover the sample efficiency and spatial learning dynamics of MLLMs under structured reasoning.
\subsection{Experimental Settings}
\label{subsec:experimental_settings}
We fine-tuned models via LoRA (\texttt{bfloat16}) on 11.2k curated CT-RATE instances, extracting consecutive Z-axis slices to preserve spatial tracking without bounding boxes. Evaluation spans four core 3D-RadVQA tasks: Anomaly Detection (T1), Image Observation (T2), Medical Computation (T3), and Existence Detection (T4). To assess cross-benchmark generalization, we further incorporate DeepchestVQA's Recognition, Visual Reasoning, and Medical Reasoning tasks.
We report standard Accuracy (\%) for T4 and DeepchestVQA. For open-ended generative tasks (T1--T3), we employ an LLM-as-a-judge to semantically parse predictions against ground-truth coordinates, computing a task-specific Accuracy (\%) that reflects genuine 3D spatial awareness rather than rigid string matching.

\subsection{Main Results and Generalization}
\label{subsec:main_results}

Table~\ref{tab:preliminary_results} summarizes performance across the 3D-RadVQA and DeepchestVQA benchmarks. Initially, 2D-pretrained baselines exhibit limited clinical reasoning (e.g., \textit{Hounsfield2D (Base)} achieves only 7.2\% in T1). Conversely, fine-tuning on \textbf{Hounsfield-CoT} unlocks massive gains: \textit{Hounsfield2D (Ours, FT)} surges to 33.5\% in T1, significantly outperforming its \textit{Lingshu} counterpart. Notably, while Accuracy spikes, the stagnant semantic overlap (BERT-F1) confirms that structured CoT supervision shifts models from surface-level textual mimicry to genuine multi-step spatial reasoning. While native 3D models like M3D establish a performance upper bound (46.0\% in T1) via dense convolutions, our data-centric approach dramatically closes this gap using purely 2D initializations, offering a sample-efficient and transparent alternative.

To rigorously evaluate the generalizability and model-agnostic impact of our paradigm, we extend our evaluation across two additional dimensions:

\begin{figure*}[t!]
    \centering
    \includegraphics[width=0.95\textwidth]{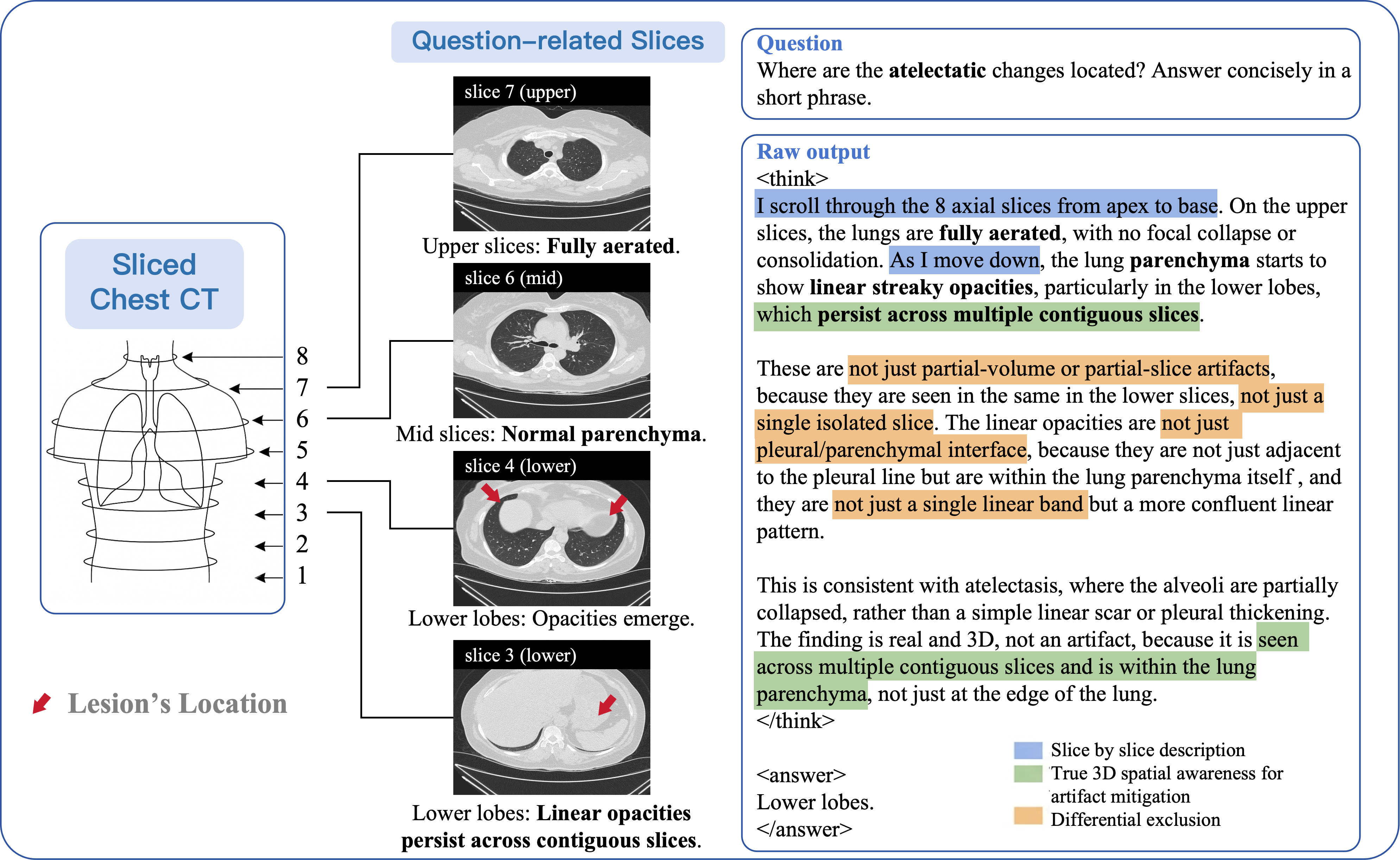}
    \caption{\textbf{Case study of Hounsfield-CoT demonstrating explicit 3D volumetric reasoning.} Given an 8-slice CT sequence for abnormality localization, the model performs explicit slice-by-slice tracking from apex to base (\textcolor{blue}{blue}). It successfully identifies linear streaky opacities in the lower lobes and leverages 3D spatial continuity (\textcolor{darkgreen}{green}) to logically rule out partial-volume artifacts and pleural interfaces (\textcolor{orange}{orange}), yielding a perfectly accurate diagnosis (BERT-F1 = 1.0).}
    \label{fig:case_study}
    \vspace{-0.5em}
\end{figure*}

\begin{itemize}
    \item \textbf{Cross-Benchmark Generalization (DeepchestVQA):} Evaluated zero-shot, \textit{Hounsfield2D (Ours, FT)} (39.6\%) outperforms robust foundation baselines like Huatuo (38.0\%) and Med3DVLM (37.9\%). While we observe minor degradation in basic \textit{Recognition} tasks—attributable to the disruption of heuristic shortcut matching—this analytical rigor translates into decisive advantages in complex categories; specifically, \textit{Medical Reasoning} improves from 39.7\% to 45.1\%.
    
    \item \textbf{Cross-Architecture Robustness (Model-Agnostic Validation):} To isolate the impact of our data paradigm from the underlying architecture, we fine-tuned the \textbf{Lingshu} baseline on our 11.2k dataset. The consistent performance surges—e.g., Organ Localization (T1) accuracy nearly doubling from 9.9\% to 19.4\%—confirm that our slice-wise data synthesis serves as a universal, model-agnostic catalyst for unlocking volumetric understanding.
\end{itemize}

\subsection{Data Scaling Ablation}
\label{subsec:data_scaling}

\begin{figure}[t!]
    \captionsetup{skip=1pt}
    \vspace{-10pt} 
    \centering
    \begin{tikzpicture}
        \begin{axis}[
            ybar=1pt, 
            width=\linewidth, 
            height=6cm, 
            enlarge x limits=0.25, 
            legend style={at={(0.5,-0.2)}, anchor=north, legend columns=3, draw=none, font=\scriptsize}, 
            ylabel={\textbf{Performance Score (\%)}}, 
            ylabel style={font=\footnotesize, yshift=-5pt}, 
            legend style={
                at={(0.5,-0.18)}, 
                anchor=north, 
                legend columns=5, 
                draw=none, 
                font=\scriptsize,
                /tikz/every even column/.append style={column sep=0.2cm} 
            },
            symbolic x coords={Task 1 (Acc), Task 2 (Acc), Task 4 (Acc)},
            xtick=data,
            nodes near coords, 
            every node near coord/.append style={font=\tiny, rotate=90, anchor=west}, 
            ymin=0, ymax=105, 
            bar width=8pt, 
        ]
        
        \addplot[fill=gray!40, draw=black!70] coordinates {
            (Task 1 (Acc), 7.2) 
            (Task 2 (Acc), 7.7) 
            (Task 4 (Acc), 77.8)
        };
        
        \addplot[fill=cyan!30, draw=black!70] coordinates {
            (Task 1 (Acc), 20.5) 
            (Task 2 (Acc), 15.1) 
            (Task 4 (Acc), 74.3)
        };
        
        \addplot[fill=blue!40, draw=black!70] coordinates {
            (Task 1 (Acc), 18.9) 
            (Task 2 (Acc), 14.9) 
            (Task 4 (Acc), 76.6)
        };
        
        \addplot[fill=blue!70, draw=black!70] coordinates {
            (Task 1 (Acc), 24.3) 
            (Task 2 (Acc), 18.0) 
            (Task 4 (Acc), 77.0)
        };
        
        \addplot[fill=blue!90!black, draw=black!70] coordinates {
            (Task 1 (Acc), 33.5) 
            (Task 2 (Acc), 24.1) 
            (Task 4 (Acc), 80.2)
        };

        \legend{Base+CoT, 25\% Data, 50\% Data, 75\% Data, 100\% Data}
        \end{axis}
    \end{tikzpicture}
    \caption{\textbf{Data Scaling Ablation on Spatial and Complex Reasoning Tasks.} We evaluate the impact of training data volume on tasks that heavily demand spatial grounding (Tasks 1 and 2) and comprehensive multi-step reasoning (Task 4). Pure computation (Task 3) is omitted here to prevent the semantic priors of the base LLM from confounding the spatial scaling trends. Spatial tasks (T1, T2) show consistent scaling improvements. Crucially, Multiple-choice Accuracy (T4) initially drops due to the structural alignment tax but ultimately recovers and peaks at 100\% scaling, proving that explicit CoT effectively replaces heuristic guessing with verified diagnostic derivation.}
    \label{fig:data_scaling_filtered}
    \vspace{-1em}
    
\end{figure}
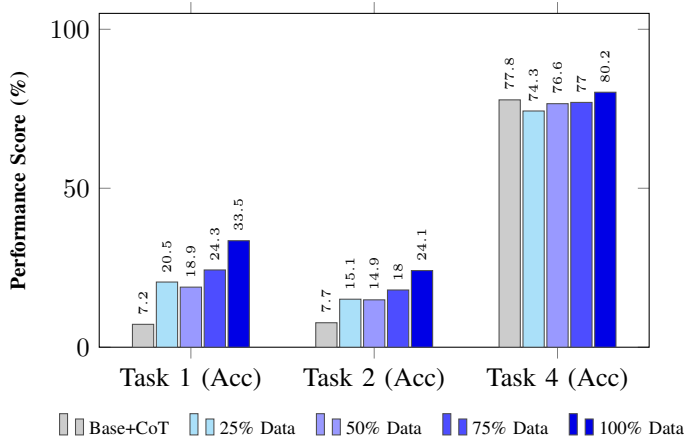

To quantitatively evaluate the sample efficiency of our structured CoT supervision, we partitioned the 11.2k dataset into four stratified subsets (25\%, 50\%, 75\%, and 100\%) while maintaining exact task ratios. We specifically selected Tasks 1, 2, and 4 from the 3D-RadVQA benchmark for this ablation, as they most directly evaluate the model's core competencies in 3D spatial grounding and evidence-based verification. As visualized in Fig.~\ref{fig:data_scaling_filtered}, the scaling reveals distinct dynamics:

\begin{itemize}
    \item \textbf{Continuous Growth in Spatial Tasks (T1 \& T2):} For Anomaly Detection (T1) and Image Observation (T2), performance scales robustly with data volume. For instance, T1 accuracy surges from a 7.2\% baseline to 33.5\% at 100\% scaling. This confirms that explicitly supervising sequential slice tracking yields direct and substantial gains in spatial awareness.
    
    \item \textbf{Shortcut Disruption in Binary Detection (T4):} Conversely, Existence Detection initially drops at 25\% data (77.8\% $\rightarrow$ 74.3\%). Unlike open-ended tasks, T4 reflects natural clinical distributions, allowing base models to inflate baselines via prior-based guessing. The limited 25\% CoT data disrupts these brittle shortcuts before genuine 3D reasoning is fully established. However, at 100\% scaling, accuracy recovers and peaks at 80.2\%, confirming that reliable visual verification has successfully replaced heuristic guessing.
    
\end{itemize}

Ultimately, this ablation demonstrates that a modestly sized, high-quality CoT dataset effectively unlocks genuine 3D medical intelligence without requiring massive data scaling.

\subsection{Case Study and Interpretability Analysis}
\label{subsec:case_study}

To qualitatively demonstrate the superior reasoning capabilities of our proposed paradigm, we present a successful case of 3D abnormality localization (Figure~\ref{fig:case_study}). Identifying atelectatic changes requires rigorous volumetric integration, a task where standard visual-language models often fail by hallucinating spatial coordinates or prematurely truncating their analysis to a single slice. 

Conversely, Hounsfield-CoT explicitly unrolls the reasoning trajectory. As illustrated in Figure~\ref{fig:case_study}, the model initiates an exhaustive volumetric traversal, scanning from the ``apex to base'', observing that the ``upper slices'' are fully aerated, and detecting ``linear streaky opacities'' exclusively in the lower lobes. Crucially, the model leverages its 3D spatial awareness to perform differential exclusion: by verifying that the opacities ``persist across multiple contiguous slices'' and reside ``within the lung parenchyma itself'', the model logically rules out single-slice partial-volume artifacts and pleural interface scarring. This evidence-based reasoning chain achieves a perfectly accurate diagnosis, demonstrating that Hounsfield-CoT effectively mirrors the systemic diagnostic workflow of radiologists, shifting MLLMs from superficial vision-to-text mapping toward genuine 3D spatial cognition.

\section{Conclusion}
\label{sec:conclusion}
In this work, we introduce \textbf{Hounsfield-CoT}, a data-centric paradigm that addresses the opaque nature of 3D medical MLLMs. By employing an Observer-Synthesizer framework, we encode the clinical slice-by-slice reading workflow into 11.2k structured reasoning instances. This approach bridges the spatial reasoning gap by synthesizing the Hounsfield-CoT data paradigm and constructing the Hounsfield3D model, which enables 2D-pretrained backbones to achieve robust volumetric reasoning and clinical interpretability. By enforcing sequential spatial tracking and differential diagnostic logic, our method effectively bypasses the need for computationally prohibitive 3D pre-training while maintaining high diagnostic fidelity.

While our paradigm yields significant gains, limitations persist. Current synthesis relies on global reports, which may occasionally lead to subtle spatial hallucinations in reasoning steps. Future work will extend this framework to diverse modalities like MRI and refine evaluation metrics to strictly verify the clinical correctness of individual thought steps. Ultimately, this research provides a transparent, scalable pathway for adapting 2D-pretrained models toward robust, high-fidelity 3D medical image understanding.

\begingroup
\footnotesize 

\bibliographystyle{unsrt}
\bibliography{main}
\endgroup

\end{document}